\documentclass{article}
\usepackage{ijcai21}

\usepackage{times}

\usepackage{soul}
\usepackage{url}
\usepackage[hidelinks]{hyperref}
\usepackage[utf8]{inputenc}
\usepackage[small]{caption}
\usepackage{graphicx}
\usepackage{amsmath}
\usepackage{booktabs}
\usepackage{floatrow}
\newfloatcommand{capbtabbox}{table}[][\FBwidth]

\usepackage{wrapfig}
\usepackage{amssymb}
\usepackage{amsmath}
\usepackage{amsmath,mathtools}
\usepackage{times}
\usepackage{caption}
\usepackage[labelformat=simple]{subcaption}
\usepackage{graphicx}
\usepackage{graphics}
\usepackage{blkarray}
\usepackage{booktabs}
\usepackage{xcolor}
\usepackage{bbold}
\usepackage{caption}
\usepackage{subcaption}
\usepackage{multirow}
\usepackage{comment}
\usepackage{epstopdf}
\usepackage{tikz}
\usepackage{todonotes}
\graphicspath{{./figures/}}
\usepackage{csquotes}
\usepackage{siunitx}
\usepackage{cite}
\usepackage{epigraph}
\usepackage{empheq}
\usepackage{comment}
\usepackage{cancel}
\usepackage{algorithm}
\usepackage{algorithmic}
\usepackage{mathtools}
\usepackage{dsfont}

\usepackage{balance}
\usepackage{dirtytalk}

\usepackage[linesnumbered,ruled,vlined,algo2e,noend]{algorithm2e}

\SetInd{0.35em}{0.5em}

\usepackage{hyperref}

\SetKwRepeat{Do}{do}{while}

\makeatother

\DeclareCaptionLabelSeparator{periodspace}{.\quad}
\newtheorem{experiment}{Experiment}

\title{A Toolchain to Design, Execute, and Monitor Robots Behaviors}

\author{
Michele Colledanchise$^1$\and
Giuseppe Cicala$^2$\and
Daniele E. Domenichelli$^{1}$\and \\
Lorenzo Natale$^1$\And
Armando Tacchella$^2$\\
\affiliations
$^1$Istituto Italiano di Tecnologia, Genova, Italy. name.surname@iit.it\\
$^2$Universit\`a degli Studi di Genova, Genova, Italy. name.surname@unige.it\\
}

\begin{document}

\maketitle

\begin{abstract}
 In this paper, we present a toolchain to design, execute, and verify
 robots behaviors following the guidelines defined by the EU H2020
 project RobMoSys. It models the robot deliberation as a Behavior Tree
 (BT) whose leaf nodes execute skills modeled as State Charts (SC)
 running as separate threads. Each state in the SC of a skill corresponds to an actuation or sensing command sent to the robot.
The toolchain provides the ability to define runtime monitors that
warn the user whenever a given system requirement is violated. We
provide a real-robot experimental validation of the toolchain and an
OS-virtualization environment to reproduce simulated experiments.
\end{abstract}

\section{Introduction}
\label{sec.introduction}
The design and maintenance of robot software are challenging as
we require it to be both reactive and modular. By reactive, we mean
the ability to quickly and efficiently react to changes; by modular,
we mean the degree to which a system’s components may be separated
into building blocks and recombined. Modularity enables components to be
developed, tested, and reused independently of one another. Furthermore,
robots may operate in open environments unattended, and the unknown
and dynamic nature of such environments may jeopardize the robot's
task execution.

This paper describes a model-based toolchain that allows a designer to
build and deploy and robot's behaviors and specification monitors for
runtime verification. The approach is based on the
framework proposed by the EU H2020 project
RobMoSys~\cite{eu2020project}. In particular,
we describe the deliberative component as a Behavior Tree (BT)
specifying a \emph{task-plot}, i.e., a sequence of
tasks required to achieve certain goals at runtime. The leaves of the
BT communicate with \emph{skills}, i.e., the coordination of
functional components made accessible to task-plots; finally, the
pieces of software that execute code at the functional layer
are \emph{components}.
We focus on runtime verification using monitors, i.e., software
components  that observe the signals exchanged by the BT, the skills
and the components and verify that they are consistent with given
requirements. We validate our toolchain in a real robot scenario and
provide an OS-virtualization environment based on Docker to reproduce
simulated experiments.

\section{Behavior Trees}
\label{sec:bg}
In this section, we briefly describe Behavior Trees (BTs), a detailed description can be found in the literature~\cite{BTBook}.
A BT is a directed rooted tree where the internal nodes are
sub-behaviors compositions and leaf nodes are either actions to be
performed or conditions to be checked. We follow the canonical nomenclature
for root, parent, and child nodes.
The children of a BT node are placed below it, as in Figure~\ref{s1.fig.complexBT}, and they are executed
 from left to right. The execution of a BT begins from the root node. It sends \emph{ticks}, which are activation signals, with a given frequency to its children. A node in the tree is executed if and only if it receives ticks. When the node no longer receives ticks, its execution stops.  The child returns to the parent a status that can be either \emph{Success}, \emph{Running}, or \emph{Failure} according to the logic of the node. Below we present the most common BT nodes and their logic.

\paragraph*{Sequence}
When a Sequence node receives ticks, it routes them to its children from left to right. It returns Failure or Running whenever a child returns respectively Failure or Running. It returns Success whenever all the children return Success. When child $i$ returns Running or Failure, the Sequence node stops sending ticks to the next child (if any) but keeps ticking all the children up to child $i$.
The Sequence node is graphically represented by a square with the label \say{$\rightarrow$}, as in Figure~\ref{s1.fig.complexBT}.

\paragraph*{Fallback}

When a Fallback node receives ticks, it routes them to its children from left to right. It returns a status of Success or Running whenever a child returns Success or Running respectively. It returns Failure whenever all the children return Failure. When child $i$ returns Running or Success, the Fallback node stops sending ticks to the next child (if any) but keeps ticking all the children up to the child $i$.
The Fallback node is represented by a square with the label \say{$?$}, as in Figure~\ref{s1.fig.complexBT}.

\paragraph*{Action}
An action performs some operations as long as it receives ticks. It returns Success upon completion and Failure if the operations cannot be completed. It returns Running otherwise. When a running Action no longer receives ticks, it aborts its execution. An Action node is graphically represented by a rectangle, as in Figure~\ref{s1.fig.complexBT}.

\paragraph*{Condition}
Whenever a Condition node receives ticks, it checks if a proposition is satisfied or not. It returns Success or Failure accordingly. A Condition is graphically represented by an ellipse, as in Figure~\ref{s1.fig.complexBT}.

\section{Toolchain Overview}
In this section, we overview the toolchain, starting from
the software abstraction layers and then delving into
the implementation details.

\subsection{Abstraction Layers}
The toolchain partitions the robotic software  into
different layers of abstractions, each addressing different concerns;
such organization enables level-specific efficient solutions~\cite{ahmad2016software}. We use the abstraction layers defined by the RobMoSys Robotic Software Component~\cite{eu2020project} that categorize the robotic software in:

\paragraph*{Mission Layer}
This layer provides the software to design the BT encoding the
deliberation policy of the robot to achieve a given goal (e.g., clean
the office). It may make use of task planners and optimization
algorithms. The toolchain does not provide nor impose any particular
solution about this layer, provided that the BT is encoded in a
specific form, as described later. In the literature, we find examples
where the BT can be synthesized either
algorithmically~\cite{colledanchise2019towards}
or manually through an API or a GUI-based
tool~\cite{BTBook}.

\paragraph*{Task Layer}
This layer encodes the BT description. We describe the BT using the
same eXtended Markup Language (XML) formalism used by
Groot\footnote{\url{https://github.com/BehaviorTree/Groot}} and
Papyrus4Robotics\footnote{\url{https://robmosys.eu/wiki-sn-03/baseline:environment_tools:papyrus4robotics}}
software libraries.

\paragraph*{Skill Layer}
This layer encodes the basic capabilities (\emph{skills}) of a robot,
e.g., grasp an object, go to a given location in the map, and the like.
The specification formalism for skills is State Charts (SC), using the
W3C's scxml formalism\footnote{\url{https://www.w3.org/TR/scxml/}}.
In our toolchain, a skill describes the implementation of a leaf node
of a BT and orchestrates a set of services
from the \emph{Service Layer}; such orchestration involves retrieving
or setting parameters, starting and stopping services, e.g., read a
pose from the map server and then send a command to the navigation
server to reach that pose.
Concerning the execution engine, skills can run in separate
executables (also separate from the BT executable), where the source
code of the skill is written in a separate component that exposes the
interface for calling the functions \texttt{Tick()} and (possibly)
\texttt{Halt()}. A  leaf node of the BT engine forwards the calls of
the functions \texttt{Tick()} and \texttt{Halt()} to the corresponding
component. In our toolchain, the YARP middleware
handles these calls over the network.

\paragraph*{Service Layer} This layer contains entities that serve as
access points to sensors and actuators of the robot. It describes the
server side of a service that performs basic capabilities (e.g., get
the battery level, move the base, etc.).


\subsection{Implementation Detail}
\label{sec:overview:implementation}
The toolchain provides two main features: $(i)$) Behavior implementation
and execution, including the execution of the BT and the SCs, and $(ii)$e
runtime monitor execution where property monitors are
modeled as SCs. As mentioned above, the BT description
can be synthesized either manually or algorithmically.
In our example, we manually design the BT using Groot.

The toolchain interprets and executes the XML of the BT using the
\emph{BehaviorTree.CPP}
engine\footnote{\url{https://www.behaviortree.dev/}} while the SCs for
the skills and the runtime monitors are executed through the \emph{Qt SCXML}
engine.\footnote{\url{https://doc.qt.io/qt-5/qtscxml-overview.html}}
We defined the communication between BT and SCs for skills
and between SCs and the components that they orchestrate using an Interface
Definition Language (IDL)  implemented via the \emph{YARP Thirft}
Remote Procedure Calls (RPCs) and handled via the YARP
middleware~\cite{metta2006yarp}, which is synchronous and follows the
RobMoSys Query communication
pattern\footnote{\url{https://robmosys.eu/wiki/modeling:metamodels:commpattern}}. Components,
on the other hand, can communicate with each other using
arbitrary communication patterns.

SCs also describe the runtime monitors. At this stage of the
toolchain, they are manually synthesized from requirements, but
algorithmic generation is feasible and inexpensive for typical
properties like safety and response. We use a feature of the
YARP middleware called \emph{portmonitor}~\cite{paikan2014data} to
intercept the messages/commands between the BT and the SCs and those
between the SCs and the components. The portmonitor then propagates
the messages to the SC of the corresponding runtime monitor.

\section{Example Scenario}
\label{sec:scenario}
In this example scenario, a service robot has to reach a predefined
location. Whenever the battery reaches below $30$\% of its full
capacity, the robot must stop and reach the charging station, where it
waits until a human operator plugs in the power cable and the battery
gets fully charged. Once the battery is recharged, the robot resumes
the previous  task.

\subsection{Behavior Tree}
\begin{figure}[h!]
\centering
\includegraphics[width=\columnwidth]{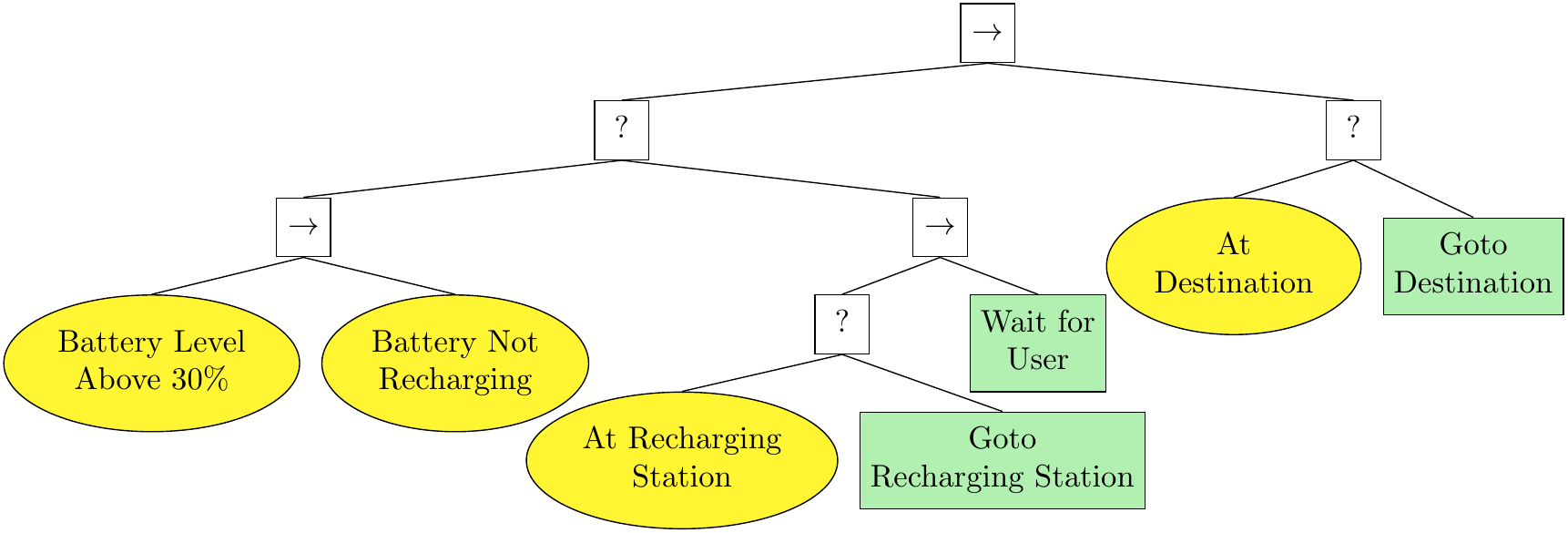}
\caption{BT for the validation scenario.}
\label{s1.fig.complexBT}
\end{figure}

The BT in Figure~\ref{s1.fig.complexBT} describes the task above.
The BT encodes the following logic:
\begin{itemize}
\item The robot checks if the battery level is below $30$\% of its capacity or if the battery is charging.
\item If the battery level is below $30$\% and it is not already under charge, the robot goes to the charging station.
\item If the battery level is charging and the robot is at the charging station, the robot waits until the battery gets fully charged.
\item If the battery level is above $30$\% and it is not under charge, it goes to the destination.
\end{itemize}

The logic of the leaf nodes of the BT above are:

\paragraph*{Battery Level Above 30\% (Condition)}
It sends a request to
a battery reader (a component in the Service Layer) that provides the battery level. The condition returns success if the level is above 30\% of its full capacity. It returns failure otherwise.

\paragraph*{Battery Not Recharging}
It sends a request to a
battery reader (a component in the Service Layer) that provides the charging status of the
battery (recharging/not recharging). The condition returns success if the battery is not
recharging. It returns failure otherwise.

\paragraph*{At Location}
It sends a request to a
localization server (a component in the Service Layer), which provides the robot's location.
The condition returns success if the robot is at the given location. It returns failure otherwise.

\paragraph*{Goto Location}
It sends a request to a
navigation server (a component in the Service Layer), to reach a predefined location, and then it waits for a result from the server (destination reached or path not found). The action
returns failure if the navigation server cannot find a path to the destination. It returns success once the
robot reaches a destination. It returns running while the robot is
navigating. If the action no longer receives ticks, it sends a request to the navigation component to stop the
robot's mobile base.

\paragraph*{Wait for User}
It returns running (to model an idle state).

\subsection{Runtime Monitors}
\label{sec:scenario:monitor}
The runtime monitors are SCs that represent behavior specifications. Currently, these SCs are manually designed. The transitions of a SC are triggered upon messages and commands passed as described in Section~\ref{sec:overview:implementation} above.

We define two specifications that the robot must fulfill:

\begin{itemize}
\item \textbf{Specification 1:} \emph{The battery level must never reach below 20\%}. That is the value read by the condition \emph{Battery Level above
  30\%} from the battery reader component must never be less than 20\%
of the battery capacity. This is because we assume that about
10\% of the battery will be sufficient to reach the recharging station
from any position.
\item \textbf{Specification 2:} \emph{Whenever the battery level reaches below 30\% of
its charge while the robot is going to its destination, the robot must
eventually go to the recharging station.} That is, the value
read by the condition \emph{Battery Level above 30\%} from the battery
reader that we should consider to verify this requirement, and then
also the commands sent by the action \emph{Goto Recharging Station} to
the navigation components that we should check.
\end{itemize}

These specifications should ensure that the logic implemented by the BT  and the component it orchestrates are correct. In Section~\ref{sec:exp} below, we show the runtime monitors for these specifications.


\section{Experimental results}
\label{sec:exp}
In this section, we present the experimental results.
We performed experiments in a simulated environment, available for reproduction.
 We also made available the video of the experiments and the code in a pre-installed OS-level virtualization environment to reproduce them.\footnote{Please visit \url{https://scope-robmosys.github.io/}}

\paragraph*{Setup}

In the simulation environment, the robot is represented with a circle and an orientation. The input of the
laser scan is computed by casting, from the center of the
robot, virtual radial beams and measuring the collision
with a point of the map defined as obstacles. We assume the absence of noise, the absence of uncertainty on the robot’s
initial position, and the absence of disturbance on
the robot's movements.

We use an Adaptive Monte Carlo Localization system to localize the robot based on the sensor input and an A$^*$-based algorithm to compute the path to the destination~\cite{randazzo2018yarp}. The simulated robot has the same software interface of the R1 robot concerning sensors and actuators. Hence, it encodes all the complexity of the real system that are relevant for the action orchestration and monitoring.

\begin{figure}[t]
\centering
\begin{subfigure}[t]{0.49\columnwidth}
\includegraphics[width=\columnwidth, trim=0 3.8cm 0 0.17cm, clip]{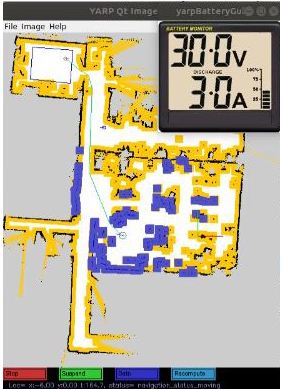}
\caption{Battery level above 30\%. The robot is reaching the destination. The safety property is satisfied.}
\label{exp1.fig.scenario1}
\end{subfigure}
\begin{subfigure}[t]{0.49\columnwidth}
\includegraphics[width=\columnwidth, trim=0 3.8cm 0 0.17cm, clip]{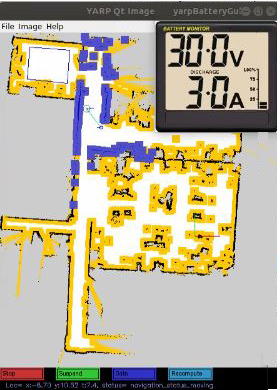}
\caption{Battery level below 20\%. The robot is reaching the recharging station. The safety property is violated.}
\label{exp1.fig.scenario2}
\end{subfigure}

\vspace*{1em}
\begin{subfigure}[t]{0.32\columnwidth}
\includegraphics[width=\columnwidth]{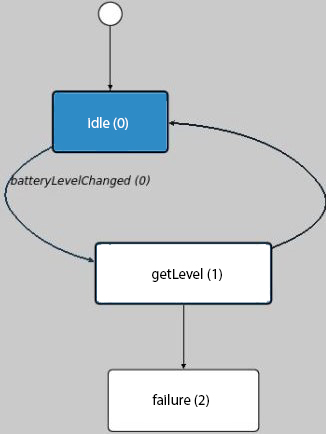}
\caption{The monitor at state Idle. Waiting for the skill to send a request to the component.}
\label{exp1.fig.exec1}
\end{subfigure}
\begin{subfigure}[t]{0.32\columnwidth}
\includegraphics[width=\columnwidth]{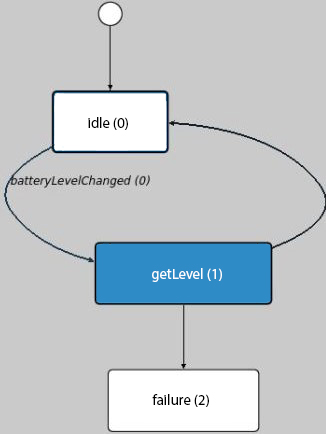}
\caption{The monitor at state Get. Waiting for the component to respond.}
\label{exp1.fig.exec2}
\end{subfigure}
\begin{subfigure}[t]{0.32\columnwidth}
\includegraphics[width=\columnwidth]{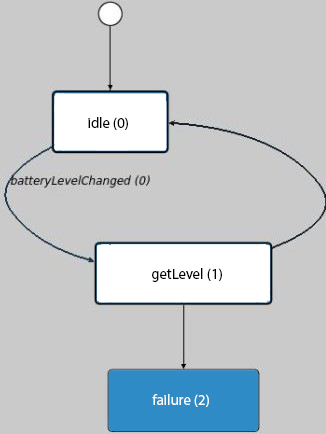}
\caption{The monitor at state Failure. Property violation detected.}
\label{exp1.fig.exec3}
\end{subfigure}
\caption{Execution steps of Experiment~\ref{experiment.battery}. The destination is the room in the top left. The charging station is the small
circle on the way to the destination. The green curve is the computed path. The blue pixels represent
objects detected by the laser scanner.}
\label{exp1.fig}
\end{figure}

\begin{experiment}[Runtime monitor for a safety property]
\label{experiment.battery}
In this experiment, we show an execution example of the runtime monitor on the battery level specification (Specification 1 in Section~\ref{sec:scenario:monitor}). The monitor implemented is the one depicted in Figure~\ref{exp1.fig.exec1}.
In particular, the monitor verifies that the property \say{The battery level must never reach below 20\%} is satisfied throughout the execution.
 To impose a violation of the property,  we manually change the battery level to 10\%.
Figure~\ref{exp1.fig} shows the execution steps of this experiment for both the scenario and the monitor.
At the initial state the battery level is above 30\% (Figure~\ref{exp1.fig.scenario1}). The BT sends ticks to the condition \say{Battery Level Above 30\%} and the condition sends a request to the corresponding skill. Then, the skill sends a request to the battery component to read the battery level. The portmonitor detects the request and sends the corresponding message to the runtime monitor and jumps to the state \emph{get} (Figure~\ref{exp1.fig.exec1}). When the component replies to the skill, the portmonitor propagates the reply to the runtime monitor. If the battery value is above 20\% the monitor moves back to the state \emph{idle} (Figure~\ref{exp1.fig.exec2}). Once we manually change the battery level to be 10\% (Figure~\ref{exp1.fig.scenario2}) the monitor detects the property violation and it jumps to the state \emph{failure} (Figure~\ref{exp1.fig.exec3}).
\end{experiment}

\begin{figure}[t]
\centering
\begin{subfigure}[t]{0.49\columnwidth}
\includegraphics[width=\columnwidth, trim=0.15cm 10cm 37cm 0.2cm, clip]{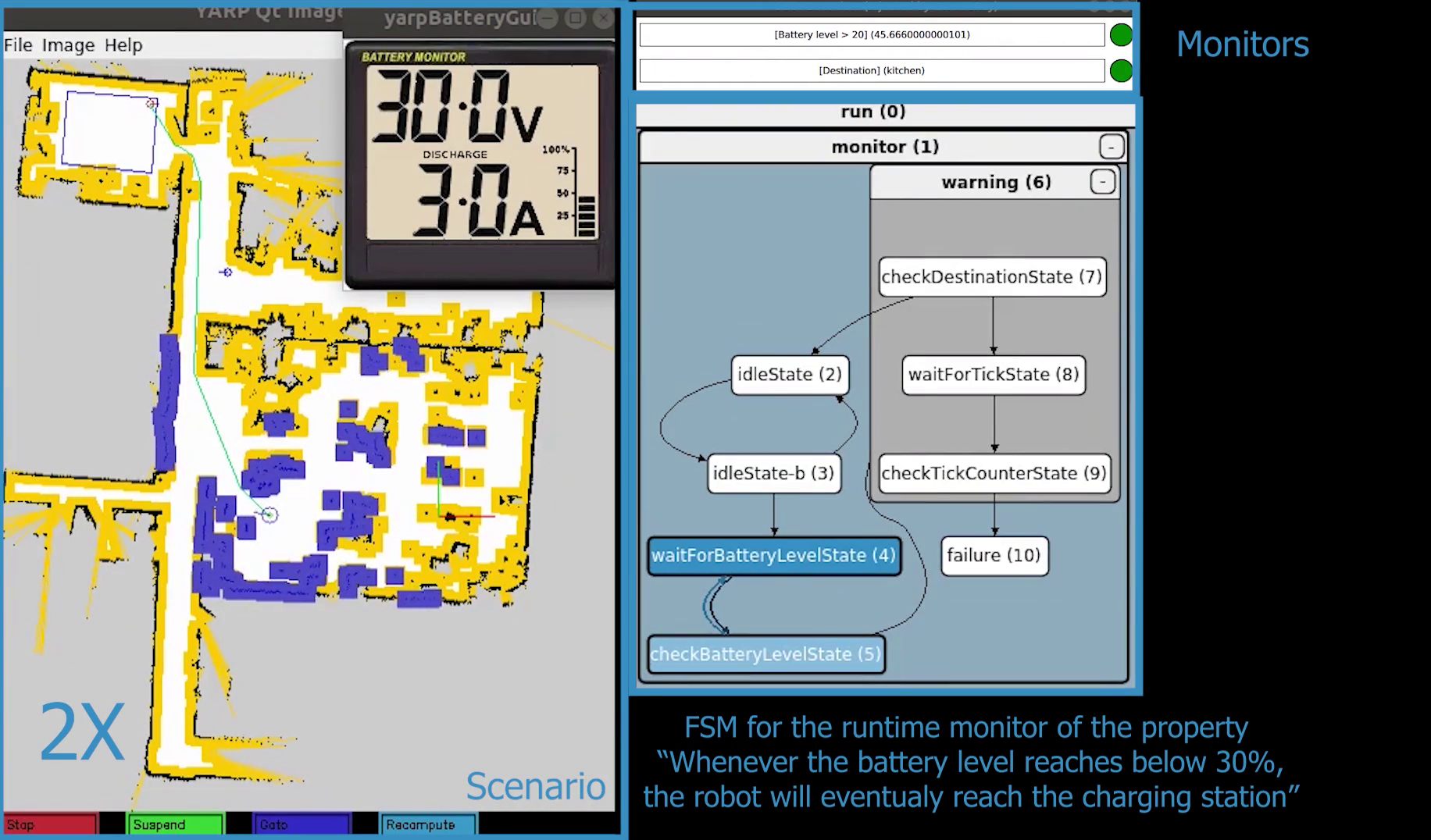}
\caption{Battery level above 30\%. The robot is reaching the destination.}
\label{exp2.fig.scenario1}
\end{subfigure}
\begin{subfigure}[t]{0.49\columnwidth}
\includegraphics[width=\columnwidth, trim=0.15cm 10cm 37cm 0.2cm, clip]{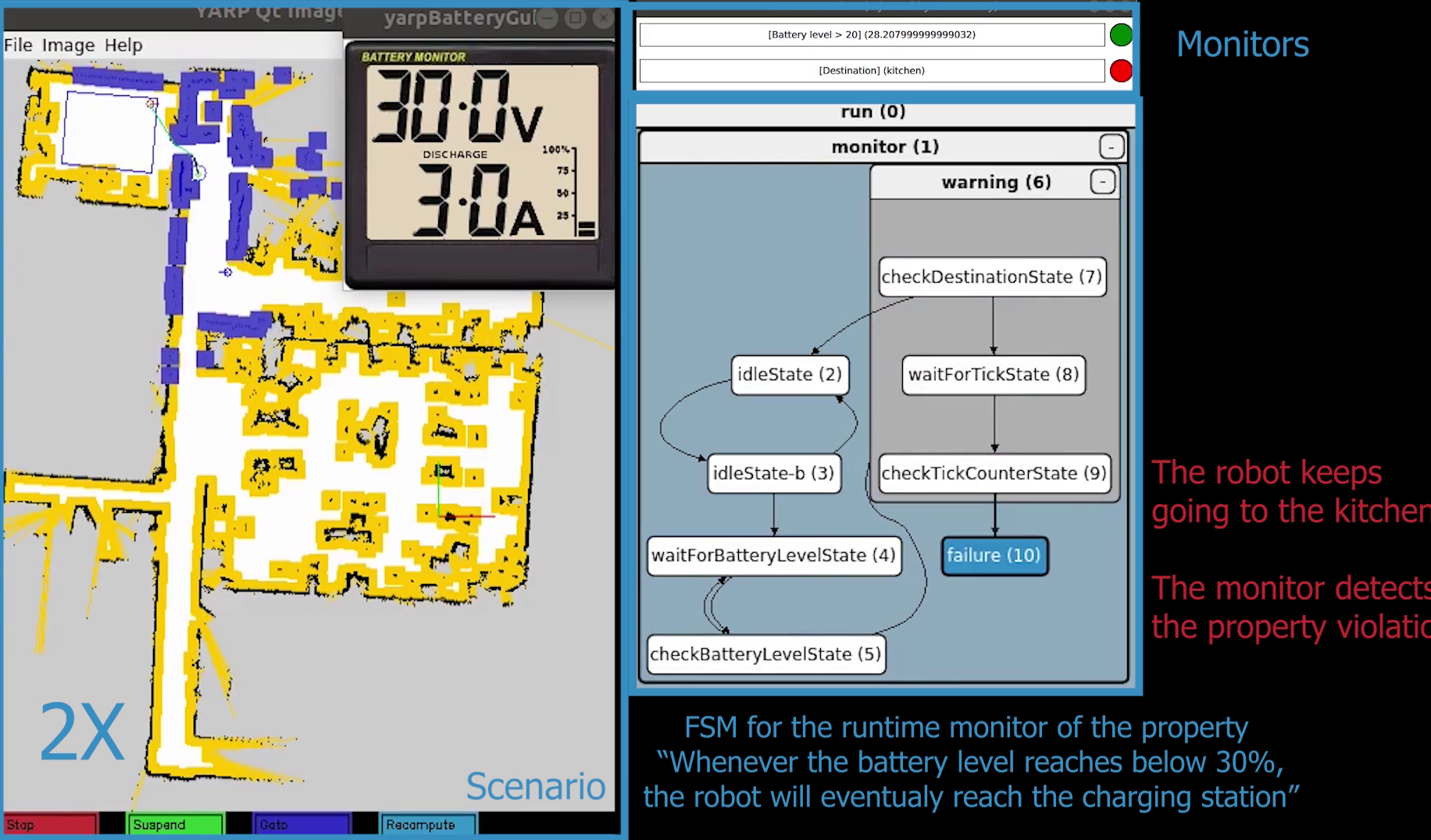}
\caption{Battery level at 25\%. The robot keeps moving to the destination. The response property is violated.}
\label{exp2.fig.scenario2}
\end{subfigure}

\vspace*{1em}
\begin{subfigure}[t]{0.45\columnwidth}
\includegraphics[width=\columnwidth, trim=29cm 7cm 13.7cm 7.4cm, clip]{exp2-s0.png}
\caption{The monitor checks that whenever the battery level is below 30\% the target  becomes the charging station.}
\label{exp2.fig.exec1}
\end{subfigure}
\begin{subfigure}[t]{0.45\columnwidth}
\includegraphics[width=\columnwidth, trim=29cm 7cm 13.7cm 7.4cm, clip]{exp2-s1.png}
\caption{ Response property violation detected.}
\label{exp2.fig.exec2}
\end{subfigure}
\caption{Execution steps of Experiment~\ref{experiment.bug}.}
\label{exp2.fig}
\end{figure}


\begin{experiment}[Runtime monitor for a  responsive property]
\label{experiment.bug}

In this experiment, we show an execution example on the runtime monitor on the battery recharging behavior specification (Specification 2 in Section~\ref{sec:scenario:monitor}). The monitor implemented verifies that the property \say{Whenever the battery level reaches below 30\% of its charge, the robot must eventually go to the recharging station} is satisfied throughout the execution.

 To impose a violation of the property, we introduce a bug in the SC of the skill \say{Battery Level Above 30\%} such that it returns success while the battery level is above 20\%.
The execution starts with the robot that plans a path to the destination and follows it (Figure~\ref{exp2.fig.scenario1}). Then, we manually set the battery level to 25\%. Due to the bug, the robot keeps moving to the destination (Figure~\ref{exp2.fig.scenario2}).
The runtime monitor checks that whenever the battery level goes below 30\% the robot eventually moves towards the charging station (Figure~\ref{exp2.fig.exec2}). It goes to an error state because the battery gets below 30\% during the robot navigation and the robot does not reach the charging station (Figure~\ref{exp2.fig.exec2}).
\end{experiment}

\section{Conclusions}
\label{sec:concl}
We presented a model-based toolchain to design, execute
and monitor robot behaviors. The toolchain follows the RobMoSys
meta-model for robot software development. It
encodes high-level behaviors in the form of a BT and basic robot
capability, in terms of actions and sensing operations, as SCs. The
clear separation of the software into different abstraction layers and
the communication interfaces defined by Thrift IDL enables an easier
design and composition of robot skills and a more effective way
to monitor the robot behaviors and detected possible specification
violations.

\bibliographystyle{abbrv}
\bibliography{refs}
\end{document}